\def\year{2018}\relax
\documentclass[letterpaper]{article} %DO NOT CHANGE THIS
\usepackage{aaai18}  %Required
\usepackage{times}  %Required
\usepackage{helvet}  %Required
\usepackage{courier}  %Required
\usepackage{url}  %Required
\usepackage{graphicx}  %Required
\usepackage{hyperref}       % hyperlinks
\usepackage{url}            % simple URL typesetting
\usepackage{booktabs}       % professional-quality tables
\usepackage{amsfonts}       % blackboard math symbols
\usepackage{nicefrac}       % compact symbols for 1/2, etc.
\usepackage{microtype}      % microtypography
\usepackage{booktabs} % For formal tables
\usepackage{graphicx}
\usepackage{multirow,url}
\usepackage{amsmath}
\usepackage{algorithm}
\usepackage[noend]{algpseudocode}
\usepackage{subcaption}
\usepackage{tikz}
\usetikzlibrary{shapes,arrows}

\makeatletter
\newcommand*{\radiobutton}{%
  \@ifstar{\@radiobutton0}{\@radiobutton1}%
}
\newcommand*{\@radiobutton}[1]{%
  \begin{tikzpicture}
    \pgfmathsetlengthmacro\radius{height("X")/2}
    \draw[radius=\radius] circle;
    \ifcase#1 \fill[radius=.6*\radius] circle;\fi
  \end{tikzpicture}%
}
\begin{document}
\title{CrowdMI: Multiple Imputation via Crowdsourcing}
\author{Lovedeep Gondara\\
Simon Fraser University\\
lgondara@sfu.ca\\
}
\maketitle
\begin{abstract}
Can humans impute missing data with similar proficiency as machines? This is the question we aim to answer in this paper. We present a novel idea of converting observations with missing data in to a survey questionnaire, which is presented to crowdworkers for completion. We replicate a multiple imputation framework by having multiple unique crowdworkers complete our questionnaire. Experimental results demonstrate that using our method, it is possible to generate valid imputations for qualitative and quantitative missing data, with results comparable to imputations generated by complex statistical models.
\end{abstract}

\section{Introduction}
Missing data is unavoidable and is a significant issue impacting all domains. Data can be missing due to a number of reasons, including but not limited to a faulty apparatus, error prone manual data entry, non response in surveys etc. Irrespective of the cause, missing data is always undesirable. Even small proportions of missing data can seriously bias inference, resulting in erroneous conclusions.

Popular and often used method for missing data is imputation. Where we replace missing values with most probable candidates. Chosen using methods from statistics or machine learning. The methods can be as simple as replacing the missing observation with the average of observed values or as complicated as modelling missing data using Bayesian models or complex models from deep learning. Instead of imputing one value for one missing observation. It is recommended and preferred to impute multiple slightly different values. This creates multiple versions of the complete dataset. It is done to induce variability in imputation process that accounts for the imputation error \cite{rubin2004multiple,schafer1999multiple}, as the true value of the missing observation is never known. The process of creating multiple versions of complete dataset is known as \textbf{multiple imputation}. Variance in multiple imputations reflect the error in imputation process. Imputed datasets are then analyzed separately and the output is combined using the methods described by Little and Rubin \cite{little2014statistical}. Multiple imputation is the most widely used framework to deal with missing data.

Computational imputation methods are limited by various factors such as the sample size of training data, missingness patterns, missing outliers, proper encoding of prior information etc. Crowdsourcing has recently emerged as a new tool for collecting data that is otherwise not available \cite{franklin2011crowddb}. Crowdsourcing missing information is a promising concept and has varying degrees of success where computational methods lack state-of-the-art \cite{brabham2008crowdsourcing,leimeister2009leveraging,goodchild2010crowdsourcing,gao2011harnessing,doan2011crowdsourcing}.

Inspired from the unique human abilities such as the large degrees of freedom, intuitive reasoning and previous success of crowdsourced problems. We propose a novel multiple imputation framework based on crowdsourcing, called CrowdMI. CrowdMI works by structuring missing data problem as a survey questionnaire, which is presented to crowdworkers for completion. This study is aimed at answering the following main questions:

\begin{enumerate}
    \item Can humans fill in (impute) missing qualitative and quantitative data?
    \item If the same missing value is imputed multiple times by humans, will it have similar variation as in values imputed by machine based multiple imputation methods?
    \item How much information is needed by crowdworkers to efficiently impute missing observations and what are the best methods to provide that information?
\end{enumerate}

This study has far reaching impact. Successful imputation of qualitative and quantitative data by humans can open research possibilities in active learning, where observations with missing data can be first imputed by humans before commencing the learning task. Leading to making use of incomplete observations, which is currently impossible. Existing crowdsourcing frameworks can leverage human computation to fill in quantitative and qualitative information in the databases using a proven systematic method.

\textbf{Our contributions} in this study are as following:
\begin{enumerate}
    \item To best of our knowledge, we present the first study to use crowdsourcing for multiple imputation.
    \item We present the novel idea of structuring missing data as a survey questionnaire, making it easier for crowdworkers to impute missing values.
    \item We present feasibility analysis of using human computation for multiple imputation.
    \item We study the impact of presenting crowdworkers with varying degrees of prior knowledge about the dataset before imputing missing values.
    \item We present the comparison of human imputed values with state-of-the-art machine imputation.
\end{enumerate}

Next section reviews some related work to missing data and crowdsourcing. Followed by the section introducing CrowdMI. Next section presents evaluation of CrowdMI and several challenges using real life datasets. Finally, we conclude the paper with limitations and directions for our future work.

This study has been approved by Office of research Ethics at XXX (Institutional Review Board Approval Number: XXX).

\section{Preliminaries}
This section provides some required background on crowdsourcing and missing data imputation.
\subsection{Crowdsourcing}
Crowdsourcing is defined as "The outsourcing of a job (typically performed by a designated agent) to a large undefined group in the form of an open call" \cite{howe2006rise,misra2014crowdsourcing}. Crowdsourcing generally refers to the idea of bringing together a group of individuals to collaborate for solving a problem. Under right circumstances, the results obtained from crowdsourcing a problem are highly reliable and the process is efficient. Now internet provides an ideal platform to implement crowdsourcing for various problems. We refer readers to Brabham \cite{brabham2008crowdsourcing} for further discussion.

\subsection{Missing data and multiple imputation}
Missing data is a well researched topic and a pervasive problem in data mining and analytical domains. As most of the statistical and machine learning models are designed to work with complete datasets only, missing data results in reduced sample size and biased inference.

Usually when dealing with missing data. A single missing value is replaced with a single predicted value, which can be chosen using a simple column average or a complex statistical model. During analysis, imputed value is treated same as all observed values. The disadvantage of doing so is obvious. Considering an imputed value same as observed values is indirectly assuming that imputation model is prefect, which is impossible in real life scenarios. To overcome this issue multiple imputation is often used.

For multiple imputation \cite{rubin2004multiple}, instead of filling in a single value per missing observation. We fill in the missing observation with multiple imputed values, each slightly different than the other. We explain this further using a simple example. Using aid of data from Table \ref{dummy_data}, representing an income questionnaire in a survey, we denote missing values with "?".
\begin{table}[]
\centering
\caption{Data snippet for income questionnaire with missing data represented using '?'}
\label{dummy_data}
\begin{tabular}{|l|l|l|l|l|l|l|}
\hline
Id & Age & Sex & Income & Postal & Job & Marital status \\ \hline
1  & 50  & M   & 100    & 123    & a   & single         \\ \hline
2  & 45  & ?   & ?      & 456    & ?   & married         \\ \hline
3  & ?   & F   & ?      & 789    & ?   & ?              \\ \hline
\end{tabular}
\end{table}
In a multiple imputation scenario, we will create multiple copies of the dataset presented in Table \ref{dummy_data} with '?' replaced by slightly different imputed values in each copy.

All imputed values are assumed to be drawn from the posterior predictive distribution of missing data, given the observed data. The goal of filling in multiple values is to capture the uncertainty of imputation model in form of variance in imputed values, which better reflects real world scenarios. Multiple imputation involves three steps:

\begin{enumerate}
    \item Fill in missing values $k$ times to create $k$ complete datasets
    \item Analyze $k$ datasets separately
    \item Combine the inference
\end{enumerate}

Point estimates such as classification accuracy and RMSE can be combined using simple average

\begin{equation}
    \bar Q = \dfrac{1}{k} \sum_{i=1}^k \hat Q_i
\end{equation}

\section{CrowdMI}\label{sec:cmi}
In this section we introduce CrowdMI, our multiple imputation model based on human computation.

\subsection{Method}
The most important challenge we face is to present available raw data to a presumably data naive crowdworker. Especially presenting it in a way so to aid the crowdworker to fill in the missing values. It is well understood that presenting rows of raw data without proper context to crowdworkers will not yield good results. Instead it will inflate the total time needed to complete the job, associated job costs, and will result in noisy answers. 

\begin{figure}[t!]
	\centering
\tikzstyle{decision} = [diamond, draw, fill=blue!20, 
    text width=4.5em, text badly centered, node distance=3cm, inner sep=0pt]
\tikzstyle{block} = [rectangle, draw, fill=blue!20, 
    text width=10em, text centered, rounded corners, minimum height=4em]
\tikzstyle{line} = [draw, -latex']
\tikzstyle{cloud} = [draw, ellipse,fill=red!20, node distance=3cm,
    minimum height=2em]
    
\begin{tikzpicture}[node distance = 2cm, auto, scale=0.7, every node/.style={scale=0.7}]
    % Place nodes
    \node [block] (init) {Get dataset};
    \node [block, below of=init] (identify) {Extract $n$ missing rows};
    \node [block, below of=identify] (evaluate) {Create survey questionnaire};
    \node [block, below of=evaluate] (create) {Post Job};
    \node [block, left of=create, node distance=4cm] (update) {Update dataset};
    \node [decision, below of=create] (decide) {Does imputed value fulfills constraints?};
    \node [block, below of=decide, node distance=3cm] (stop) {stop};
    % Draw edges
    \path [line] (init) -- (identify);
    \path [line] (identify) -- (evaluate);
    \path [line] (evaluate) -- (create);
    \path [line] (create) -- (decide);
    \path [line] (decide) -| node [near start] {yes} (update);
    \path [line] (update) |- (identify);
    \path [line] (decide) -- node {no}(stop);
\end{tikzpicture}

\caption{CrowdMI mechanism of action: $n$ missing data rows are extracted from the candidate dataset with missing data and are converted into a survey questionnaire providing details about the dataset and extra available information (varying degrees). The questionnaire is then answered by human participants using a crowdsourcing framework, answers are treated as imputed values and the dataset is updated. Process is repeated $k$ times to get multiple imputations.}\label{processimg}
\end{figure}
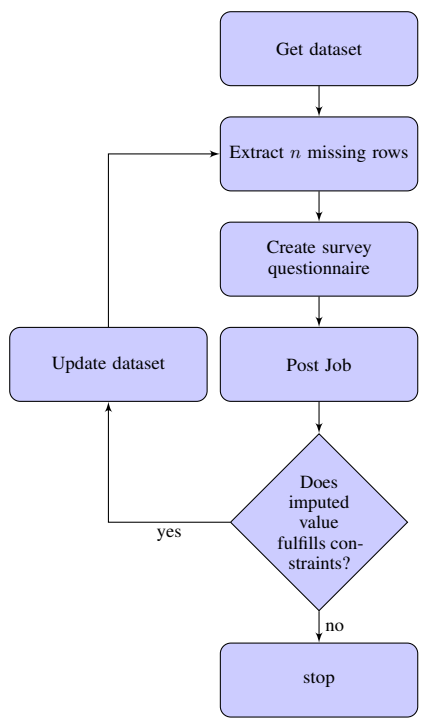

Surveys and questionnaires are known to be one of the best methods to extract crowd knowledge \cite{de2005mix,wright2005researching} in a structured format using indirect communication. We chose the same route, a survey questionnaire is designed that presents some preliminary information about the dataset and then each row with missing data is presented as an easily interpretable survey question. In a basic format, our process is presented using Figure \ref{processimg}.

From a theoretical perspective, missing data imputation using crowd can be considered as a special case of active learning. Where the survey responses to fill in the missing data are synonymous to the labels provided by users in active learning frameworks.

Considering the widely used strategy of uncertainty sample in active learning, where we select a sample for query, that we are most uncertain about \cite{lewis1994sequential}. We can immediately see that in case of missing data, we will choose rows with missing observations to get information so they can be used in further analysis. We refer readers to the work of Yan et al. \cite{yan2011active} and Fang et al. \cite{fang2014active} for further details on the role of crowdsourcing in active learning.

It is well known that the quality of survey data is directly proportional to respondents capacity and willingness to provide accurate and reliable answers \cite{sanchez1992effects}. Uninteresting questionnaires coupled with the length and complexity tend to get more noisy answers compared to a well designed questionnaire with intuitive directions and a limited number of questions. Hence, we follow some basic design principles outlined below.

\begin{enumerate}
    \item \textbf{Keep questionnaire short}: Long questionnaires are known to cause fatigue in participants. So, we limit the maximum number of questions to 10. But, this will only allow us to get 10 imputations. A natural question is: what if we have more than 10 missing observations? Our design is easily extended by creating different questionnaires with different missing rows and releasing them to a crowdsourcing platform as individual questionnaires, results of which can be combined later to get a complete dataset.
    
    \item \textbf{Keep questionnaire simple}: This is arguably the most important design choice in our case. As our datasets can be from different domains, such as medical informatics, sales etc. Crowdworkers are not assumed to have any prior knowledge. Using simple language to phrase the information about the dataset and the survey questions themselves is vital. If we use technical jargons avoiding explanation of concepts in a layman's context, we risk inferior quality responses. We need to have a balance in the complexity of the language used and the information conveyed for best possible results.
    
    \item \textbf{Use visual representations}: Long and tedious text only or numbers only descriptions induce boredom and contribute to random noisy answers by crowdworkers. To provide visual stimulation with information, we decided to use plots showing the data distribution with respect to the attribute with missing values.
    
    \item \textbf{Follow basic design aesthetics}: Surprisingly, the aesthetics play a significant role in questionnaire response and the quality of responses received. Keeping the basic design principles in mind, we use easy to read font with black color on white background while avoiding unnecessary use of colors.
    
\end{enumerate}

\subsection{An example}
We demonstrate the questionnaire generation process by an example using a sample dataset shown in Table \ref{galton}.

\begin{table}[h!]
\centering
\caption{Sample data from Galton's height dataset. Father and mother column hold the information for height of subject's parents, with gender holding subject's sex information and height has the information for the subject's height.}
\label{galton}
\begin{tabular}{llll}
Father & Mother & Gender & Height \\
78.5   & 67     & M      & 73.2   \\
75.5   & 65.5   & M      & 73.5   \\
75     & 64     & F      & 68     \\
69.5   & 64.5   & F      & 63.7  
\end{tabular}
\end{table}

We start by providing context around the dataset as an introduction to the survey. In this particular case, the tuples belong to the famous \emph{Galton's height data}. Which investigates the relationship between offspring's and their parent's height. A sample survey introduction based on the dataset is framed as following:

"This dataset presents relationship between the heights of parents and their adult children. It is seen that there a positive relationship between parent's and their offspring's height. In other words, taller parents tend to have taller children. It is also observed that on average males are taller than females."

Additional details such as attribute distribution or relationship between different attributes can be provided if needed, either in forms of tables or as plots. After providing the introduction and other necessary details, a survey question is formulated as following:

\vspace{0.5cm}
\textbf{Q}: We have a data record with missing height information for a child. Given that the gender is male, height of the father is 78.5 inch and mother is 67 inch. What do you think is the most probable height for the child?

\textbf{A}: \underline{\hspace{3cm}}
\vspace{0.5cm}

To mimic multiple imputation, same question is asked to \emph{k} participants. Which completes one iteration of our multiple imputation model for a single row of missing data. $k$ here is the parameter chosen by the end user. It dictates the number of imputations required for a single missing observation. Multiple rows of missing data can be combined as part of a single survey, called a \emph{batch} and several \emph{batches} can be combined into a single \emph{job}.

An obvious question is, "How many imputations are needed?". That is, what value of $k$ should we use? This is a crucial decision, as it will impact the financial decision making for use of crowdsourcing in a missing data scenario. Our primary goal, similar to multiple imputation, is to draw imputations at random from posterior predictive distribution of missing data, given observed data. In a way as to provide enough variation in the imputed values. There is no strict rule of thumb for this decision, however statistical simulations have shown \cite{graham2007many,von2005teacher} that imputations ranging from 5 to 100 can be needed depending on proportion of missing data. That is, higher the proportion of missing data, more imputations are needed. However, about 10 imputations are sufficient for good results. Which in our case, will keep crowdsource budget at a minimum.

\section{Evaluation} \label{sec:eval}
This section presents feasibility analysis and evaluation of CrowdMI on real life datasets under varying conditions.

\subsection{Datasets}
We begin with an introduction to the datasets used for evaluating CrowdMI. We use publicly available datasets that explore the relationship between respiratory function(measured using Forced Expiratory Volume(FEV)) and smoking, and diabetes in females of at least 21 years of age of Pima Indian heritage \cite{tager1979effect,smith1988using}. 

In the \emph{first dataset}, FEV is the amount of air an individual can exhale in first second of forceful breath. Data includes measurements on FEV(litres), age(years), height(inches), gender(male/female) and smoke(yes/no). First few rows of dataset are shown in Table \ref{fev} for better understanding.

\begin{table}[h!]
\centering
\caption{FEV data, first column holds the age of the subject, second column holds the information for FEV, third and fourth columns show information for height and gender of the subject and last column shows smoking status of the subject.}
\label{fev}
\begin{tabular}{lllll}
Age & FEV & Height & Gender & Smoke \\
9   & 1.708 & 57.0 & F & No  \\
8   & 1.724 & 67.5 & F & No\\
7   & 1.720 & 54.5 & F & No   \\
9   & 1.558 & 53.0 & M & No
\end{tabular}
\end{table}

For \emph{second dataset} related to diabetes in females of at least 21 years of age of the Pima Indian heritage. We have information on number of times a subject was pregnant, Plasma glucose concentration,  Diastolic blood pressure, Triceps skin fold thickness, 2-Hour serum insulin, Body mass index, Diabetes pedigree function, Age and an indicator for diabetic status. First few rows of this dataset are shown in Table \ref{pima}

\begin{table}[h!]
\centering
\caption{First few rows of dataset related to diabetes in females of at least 21 years of age of Pima Indian heritage. Pg is the number of times a subject was pregnant, GL is plasma glucose concentration,  BP is diastolic blood pressure, Tr is Triceps skin fold thickness, In is 2-Hour serum insulin, Ms is body mass index, Pr is diabetes pedigree function, Ag is age and Ds is an indicator for diabetic status. It is clear that this dataset is much more richer in information compared to the FEV dataset.}
\label{pima}
\begin{tabular}{lllllllll}
Pg & Gl & BP & Tr & In & Ms & Pr & Ag & Ds \\
6   & 148 & 72 & 35 & 0  &33.6 & 0.63 & 50 & Pos\\
1   & 85 & 66 & 29 & 0 & 26.6 & 0.35 & 31 & Neg\\
8   & 183 & 64 & 0 & 0 & 23.3 & 0.67 & 32 & Pos \\
1   & 89 & 66 & 23 & 94 & 28.1 & 0.17 & 21 & Neg
\end{tabular}
\end{table}

To create a \emph{third dataset}, we use FEV data and perturb some of the rows. This is done in order to create a new questionnaire, results of which will demonstrate the quality of responses received and how do crowdworkers react to small changes in survey questions.

\subsection{Setup}
As it is clear from the previous section, we are using datasets containing clinical information. The datasets do not have simple and intuitive information and explanation similar to the offspring's height dataset used as an example in earlier sections. 

We decided to use these datasets on purpose. As we assume that most of our crowdworkers are data naive, or naive enough not to understand the mechanics behind several medical phenomenons and their relationship with other attributes. This gives us the opportunity to design the perfect setup for our novel study, as if we do get good results on these datasets. We can assume better performance on relatively \emph{easy} datasets, encountered in day to day scenarios.

We start our experiments using the first dataset FEV by setting ten observations to have missing values for age and gender at random and pose the imputation problem as a survey question to crowdworkers. We chose age and gender to get an idea of imputation process with different data types, as age is a continuous measure and gender is binary. CrowdFlower is used as a crowdsourcing platform in all of our experiments.

\subsection{Model calibration}
Our first challenge is to calibrate our imputation model. That is to setup the model for optimal results. For proper calibration, we need to understand the type of responses we receive and amount of attention a crowdworker pays to the survey details. So for our test job, we created a simple questionnaire with a basic data description, stated as following:

"In a data set, we have subjects with age from 3 to 19 years old, with half over 10 years of age. We also have about 51\% males and 49\% females. We have a case that has age and gender missing. Based on the information provided please fill in the values you think are most probable."

Radio buttons were provided to the crowdworkers to select gender and a free text field to enter values for age. Under the free text field, we provided a help text reminding crowdworkers of the valid range for age. The survey was run with 100 \emph{judgments}. A judgment is defined by CrowdFlower as a valid answer. Each judgment has to be completed by a unique user to prevent same users from completing our survey multiple times. We set no constraints over user ability. That is all categories of users from beginner to expert were able to complete this initial survey. Initial compensation of \$0.1 per judgment was used. Our expectation was to get a uniform distribution for age within our predefined age limit and gender distribution similar to the one described in initial dataset description. 

Main questionnaire was preceded by an instruction to read descriptions carefully and ending with asking for the contributor's ID. Contributor ID is asked to enforce quality control as is recommended by CrowdFlower itself.

\begin{figure}[h!]
	\centering
		\centerline{\includegraphics[scale=0.3]{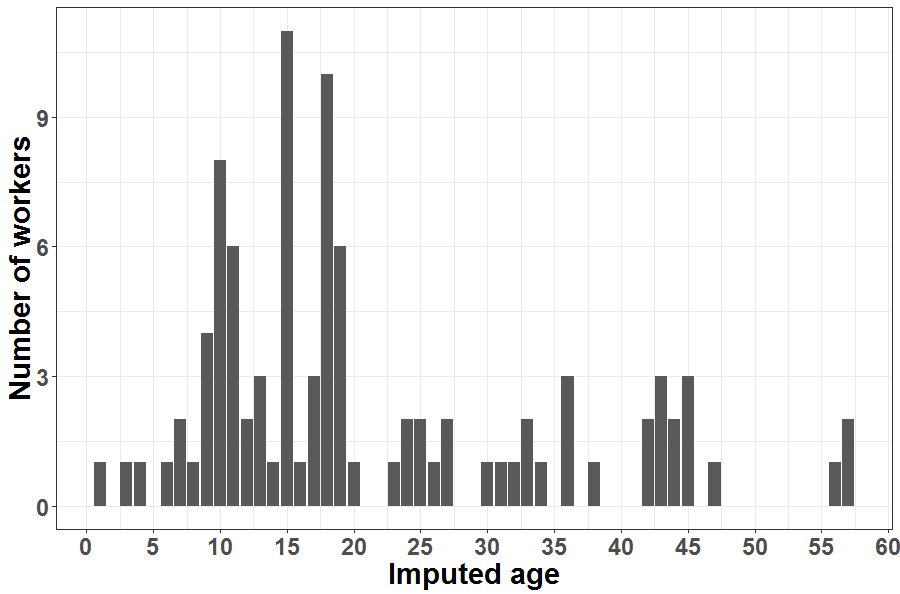}}
		\caption{Age distribution from initial imputation. Results show that crowdworkers did not adhere to the guidelines or the data description.}
		\label{job1ageugly}
\end{figure}

Initial results were disappointing. Returned age distribution was out of our predefined age limit and gender distribution was biased towards males with 83\% imputed as males and 17\% as females. Initial results of this survey are shown in Figure \ref{job1ageugly}.

These results show that crowdworkers do not adhere to guidelines or suggestions and try to complete the task as soon as possible, a confirmation of the findings in many previous crowdsourcing studies. The poor performance in our case can also be attributed to no control over contributor's experience level. As a pool of workers without experience can have a net negative impact on imputations. 

To answer some of the questions, and to measure the impact of constraints on range of crowdworkers input. We reran the same experiment, but this time using a conditional constraint where users were only allowed to enter age from 3 to 19 years. Hence forcing them to stay within the limits of our age distribution. Results were better compared to the last run, with the imputation distribution centered around the maximum allowed age value. Imputation proportion for "female" gender increased by 7\%, that is, 17\% to  24\%. 

To measure the effect of a crowdworker's experience on imputation process. We added an additional constraint where only crowdworkers at highest experience level were allowed to complete the survey. Results were improved further. Using most experienced batch of crowdworkers, age distribution was more closer to empirical distribution with 34\% less than or equal to 10. Imputed gender distribution improved as well, with 32\% imputations for female. 

So, for the rest of our experiments, we decided to use most experienced crowd with compensation fixed at \$0.15 per judgment. We restrict judgments for each question at 30, which means 300 total judgments for 10 missing values.

\subsection{Information selection}
Next significant challenge is to figure out what type and what amount of information is required by the crowdworkers to facilitate better quality responses. What criteria should be used to select the attributes that are related to the missing data and how the information should be framed as an easy to read survey description. We decided to use a simple approach of descriptive statistics. Where we investigate the relationship of the missing attribute with the rest of the attributes in the dataset and limit the information provided to the attributes that have strongest correlation with the missing attribute. This is in principle similar to how regression based imputation works.

Using dataset number one (FEV), and setting ten random rows to have missing age values. We start with the basic description of inter-variable relationships. We structure our survey introduction as following:

"This data concerns FEV (Forced Expiratory Volume), a measure of lung functionality in participants 3 to 19 years of age. We know that:
\begin{itemize}
    \item FEV increases with age and height
    \item Minimum and maximum FEV in our case is 0.79 and 5.79
    \item For a 5 year old, average FEV is 1.6 and for a 10 year old, average FEV is 2.7.
    \item A participant between heights of 55 and 60 inch have average FEV of 2.0 and participants taller than 70 inches have average FEV of 4.3
    \item Females have slightly lower FEV than males which averages at 2.5 in females compared to 2.8 in males."
\end{itemize}

\begin{figure} [h]
\centering
   \begin{subfigure}[b]{0.38\textwidth}
   \includegraphics[width=1\linewidth]{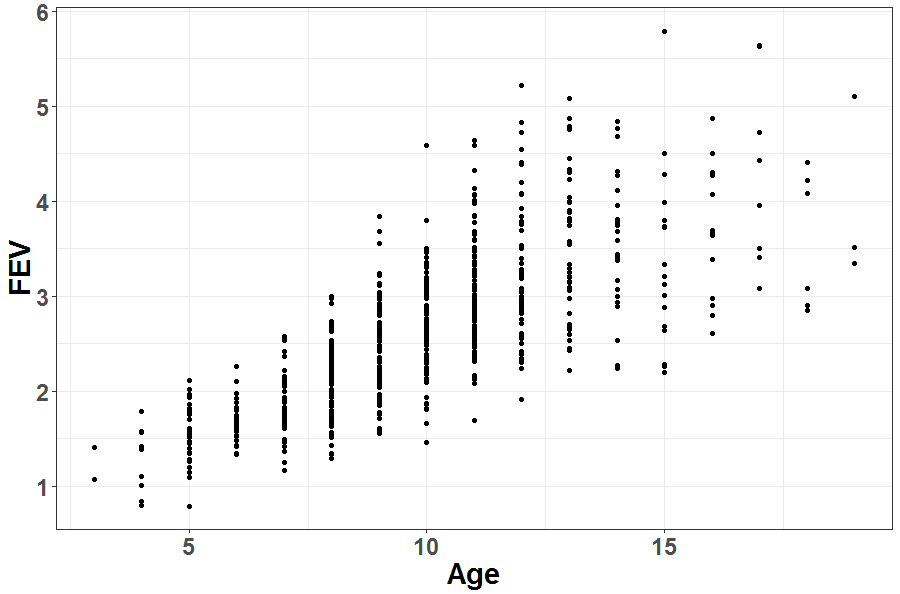}
   \caption{}
   \label{fig:Ng1} 
\end{subfigure}

\begin{subfigure}[b]{0.38\textwidth}
   \includegraphics[width=1\linewidth]{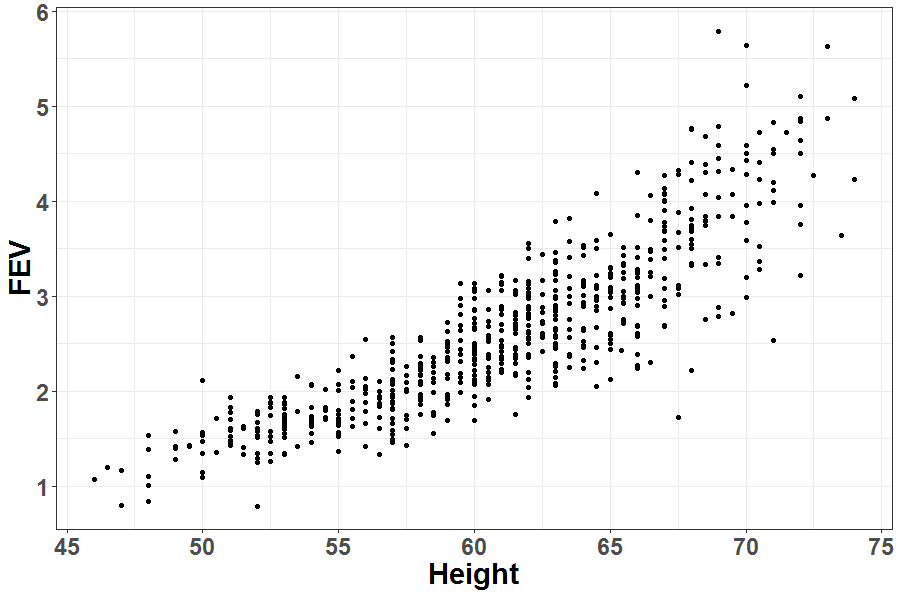}
   \caption{}
   \label{fig:Ng2}
\end{subfigure}

\caption{Scatter plots showing the relationship between the attribute with missing values and other attributes in the dataset. These are presented to the crowdworkers for visual aid to impute missing values.}\label{extra_info}
\end{figure}

We also provided supplemental information in form of two scatter plots, shown in Figure \ref{extra_info}.

This is to be noted that we did not provide descriptives that are specific to a missing attribute. But instead have given a general overview of the dataset that includes the relation of missing attribute with other variables. This is done on purpose to estimate the crowd capacity to reason relationships between different variables. So to generalize this method for cases when more than one variables are missing. After providing the information, we presented our questionnaire to crowdworkers, with one question per line and designed following the principles discussed in the earlier section.

\subsection{Comparison with machine imputation}
To get an estimate of how good CrowdMI imputations are, we need a competitor. One of the state-of-the-art methods in multiple imputation is Multiple Imputation by Chained Equation (MICE) \cite{buuren2011mice}, built on the top of Predictive Mean Matching (PMM) \cite{rubin1986statistical,little1988missing}. It is a complex statistical model and works as following: Given some variable $x$ with missingness and a set of complete variables $z$, the model works by regressing $x$ on $z$ producing regression coefficients $\beta$. Random draws are then made from posterior predictive distribution of $\beta$'s, creating a new set of coefficients $\beta^*$. Using $\beta^*$, the model then generates predictive values for $x$ for all cases and for cases where $x$ is missing. It chooses a set of cases with observed $x$ whose predicted value for $x$ is closer to predicted values for $x$ with missingness. Then from the chosen $k$ \emph{close} cases, it randomly selects one and uses it as an imputed value. To get comparable results with CrowdMI, we ran MICE based imputation to impute 30 datasets as part of multiple imputation.

This comparison is made interesting by the fact that MICE has access to all of the raw data and can model complex associations whereas the crowd workers only have information on the general statistics of the missing values coupled with what information we decide to share with them. We chose this state-of-the-art imputation model for one main reason. If crowdworkers can perform at par with MICE for imputation tasks, then there is an obvious research direction and advantage in missing data scenarios where it is better to use crowdsourcing to gather missing data compared to one shot machine based models.

Imputation results from crowd imputed data and MICE along with original values are shown in Table \ref{cimp1}. As the results are from multiple imputations (30 imputations), they are displayed using median imputed value with 25th and 75th percentile of the distribution of imputed values.

\begin{table}[h!]
\centering
\caption{Imputation results for missing age, results are shown using the median and the 25th and 75th percentiles of imputations for an observation. Results show that CrowdMI imputes values much closer to the original value compared to the imputations by MICE.}\label{cimp1}
\label{imp1}
\begin{tabular}{lll}
Original & CrowdMI & MICE\\
5   & 6.0(5.0,7.0)  & 6.0(6.0,6.8)\\
10   & 11.0(8.0,12.0) & 12.0(11.0,12.8)\\
10   & 10.5(8.0,12.0) &  12.5(11.0,17.3)\\
11   & 11.5(9.3,13.0) & 8.5(8.0,9.0)\\
10 & 12.0(10.0,13.0) & 12.0(12.0,15.5)\\
12 & 13.0(12.0,14.8) & 9.0(8.0,9.0)\\
11 & 14.0(12.0,15.0) & 11.5(11.0,12.0)\\
14 & 14.0(11.5,16.0) & 12.0(11.0,14.8)\\
14 & 16.0(13.0,16.0) & 12.0(11.0,13.0)\\
16 & 16.0(13.3,17.0) & 12.5(11.0,13.8)

\end{tabular}
\end{table}

Results show that almost all CrowdMI imputations cover missing data distribution with some median imputed values exactly same as the missing observations. CrowdMI imputations are more impressive and  less variable as compared to imputations from MICE. Which sometimes did not cover missing value between 25th and 75th percentile of all imputations.

These results are very encouraging. Hence, as of our next step, we are interested to investigate if the displayed plots provided any additional assistance to crowdworkers. As reading plots and making correct inference is not an easy task for a data naive person and visuals can be distracting from the contextual information. Hence, we again administered the same survey. But this time after removing the plots. Distribution of results was similar but coverage of original missing values was worse compared to the results obtained with the use of plots. This shows that an effective visualization of data plays a vital role in a good design of human powered imputation framework.

To challenge crowdworkers further, we decided to create a little more complicated scenario. This time, we randomly set 10 observations to have missing values for gender. We did not provide any additional information to crowdworkers compared to the last designed questionnaire. But, we added another plot displaying empirical gender distribution and FEV in the dataset.

We framed our questions as following with radio buttons provided for input.\\
\textbf{Q}:"What is the gender given that FEV is 2.4, height is 62.5, and age is 11?
\begin{itemize}
\item[\radiobutton] Male
\item[\radiobutton] Female
\end{itemize}

We would like to emphasize that imputing a categorical variable is far more error prone compared to a continuous value. Because there is no definition of an error margin or an acceptable range. The answer is either "True" or "False". We administered the survey for 10 questions with 30 judgments each. To get a comparison with machine imputations, we again used MICE with 30 imputations.

Table \ref{cimp2} shows the results. It shows that distribution of imputed values using CrowdMI is very similar to the ones imputed by MICE. Both models agreed on all results but one, where MICE imputed the correct value and CrowdMI got it wrong by two votes.

\begin{table}[h!]
\centering
\caption{Imputation results for missing gender. First column shows original value with second and third showing imputed values by CrowdMI and MICE. Values in CrowdMI and MICE show the raw number of votes received by each category from 30 imputations. Both models agree on all observations but one. }
\label{cimp2}
\begin{tabular}{lll}
Original  & CrowdMI(Male - Female) & MICE (Male - Female)\\
Female  & 13 - 17 &  13 - 17\\
Male  &   27 - 3 &    20 - 10\\
Female  &  24 - 6 &  25 - 5\\
Male  &   5 - 25 &    15 - 15\\
Female  & 28 - 2 &  19 - 11\\
Female  & 4 - 26 &  12 - 18\\
Female  & 17 - 13 &  10 - 20\\
Male  &   25 - 5 &    19 - 11\\
Male  &   28 - 2 &    17 - 13\\
Female  &  14 - 16 &  8 - 22

\end{tabular}
\end{table}

\subsection{Perturbed data}
For this comparison, we use the systematically perturbed version of FEV dataset (dataset number three). Perturbation is performed within the age attribute where original age values are replaced by new smaller age values. Now, the resulting survey is not an exact copy of first survey. But, it is also not a completely new survey. 

Running the survey for same number of iterations (30). This time imputation resulted in an overall increased proportion of gender being selected as "male" (about 57\% imputed values were male compared to near 40\% in last survey). This increase in selected male proportion is attributed to the perturbation where we decreased age from original value, keeping height the same. Which weighs the decision of crowdworkers towards selecting "male" for missing observations. This adds to our claim of good imputation quality where crowdworkers pay careful attention to details.

\subsection{A more complicated scenario}
To test CrowdMI on a more complex task. We use dataset number two (Pima Indian diabetes). This dataset is much more richer in information compared to FEV dataset. Similar to previous runs, we randomly set ten observations to have missing values for diabetic status and frame the survey questionnaire as following:

"This data has information on diabetes status for females aged 21 years and older. We know that:
\begin{itemize}
\item Diabetes positive patients have higher blood glucose levels.
\item Glucose levels are at average of around 100 for diabetes negative and 130 for diabetes positive patients.
\item Diabetes positive patients also have slightly higher blood pressure, which is about 70 compared to 65 for diabetes negative 
\item Blood pressure is also higher in people with higher body mass index.
\item Blood pressure is around 77 for people with mass greater than 40 and around 70 for mass less than 40.
\item Diabetes positive are also of older age compared to diabetes negatives patients, that is, average age for diabetic is 35 compared to about 25 for non diabetic.
\item Diabetics also have on average lower insulin compared to negatives, that is 68 compared to 100 in negatives."
\end{itemize}

In addition to the introduction above, we also provided two box plots to show the relationship between diabetes status and age, and blood glucose and diabetes status. Factors that can be easily understood by naive audience.

\begin{table}[h!]
\centering
\caption{Imputation results for diabetic status, first column shows original values in the dataset Second column is CrowdMI results and third column is results from MICE. Pos is positive, Neg is negative. Values in CrowdMI and MICE show the raw number of votes received by each category from 30 imputations. Imputation distribution is similar using CrowdMI and MICE, they disagree on two results.}
\label{imp3}
\begin{tabular}{lll}
Original  & CrowdMI(Pos - Neg) & MICE(Pos - Neg)\\
Pos  & 17 - 13 & 22 - 8\\
Pos  & 25 - 5 & 21 - 9\\
Neg  & 5 - 25 & 6 - 24\\
Pos  & 22 - 8 & 22 - 8\\
Neg  & 5 - 25 & 2 - 28\\
Neg & 4 - 26 & 4 - 26\\
Pos  & 21 - 9 & 25 - 5\\
Pos  & 5 - 25 & 18 - 12\\
Neg  & 21 - 9 & 12 - 18\\
Pos  &8 - 22 & 11 - 19

\end{tabular}
\end{table}

Table \ref{imp3} shows the results. CrowdMI and MICE disagree on two results, where MICE imputed the correct value. Although the difference was small in one misclassification, we do believe this is due to the fact that MICE has access to all available data whereas CrowdMI only uses what we decide to share with crowdworkers, hence giving MICE an unfair advantage in this scenario. Giving more contextual information should improve the outcomes with crowd.

\subsection{Participant satisfaction}
Using the results of the default survey administered by CrowdFlower to gather crowd responses for quality, ease of work, and compensation. Our surveys ranged from "easy" to "average" for question difficulty, meaning that we have room to provide more information for improvement of our outcome. Compensation is deemed "fair" by crowdworkers. Instructions for survey completion are reported as "clear". Even though we use most experienced crowd with reasonably difficult questions, we had a net "positive" response with over 90\% judgments received within first 3 hours of launching the survey.

\subsection{Is human mind Bayesian?}
In this section we try to evaluate if given extra information that is not already contained in the dataset. Can we influence crowd decision? This is in spirit similar to Bayesian methods where we use prior knowledge to update posterior probabilities. This can be useful in various real life scenarios, where researchers already know or can make strong assumptions about the missing data distribution. 

Influencing crowd imputation using prior knowledge can avoid inconsistencies due to small sample size or disproportionate sampling. An example can be of a small sample with missing gender information, where the sample is from a breast cancer study. Investigator knows that breast cancer affects majority of females but also a minority of males. If not providing any information or beliefs, crowdworkers can impute missing gender as all females or with an equal proportion of males/females. Whereas, if the investigator states explicitly that in this scenario we can expect to see 10\% of males with breast cancer. This information can nudge crowdworkers with a certain probability to impute males versus all female imputation.

In order to do evaluate this, we use the first dataset (FEV). We added a little blurb at end of our questionnaire

"However, in addition to the information contained in our dataset, we also know that in general population related to this study, females account for about 65\% of total."

Results were not significantly different from previous observations, but there was a noticeable shift, with reduced male imputation proportion. We strongly believe that crowdworkers can be further influenced by placing the extra information at start of the questionnaire compared to the end. As most crowdworkers might not pay full attention to the text below the rest of description.

\section{Related work}
\subsection{Missing data}
Most common and easiest to use method to deal with missing data is \emph{complete case analysis} in which any instance with missing data is deleted from final analysis. Filling in plausible values using mean or most frequent label from observed data is also common. There are number of studies in statistics and machine learning \cite{stephens2005accounting,efron1994missing,yuan2010multiple,jerez2010missing,allison2001missing} that propose advanced  methods for missing data imputation.

\subsection{Crowdsourced data completion}
Very few studies in crowdsourcing have attempted to leverage human computation for missing data. CrowdDB \cite{franklin2011crowddb} uses crowdsourced queries, which can include searching and filling in missing values such as an address or email for a person. Ye et al. \cite{ye2014capture} proposed a human-machine hybrid approach, involving a model imputing multiple missing values and an oracle(human) selecting the best fit. This severely limits human degrees of freedom, which in turn will limit the model's performance. Analyzing all imputations as a machine based multiple imputation model is a better alternative in such scenarios. Park and Widom \cite{park2014crowdfill} proposed Crowdfill that collects structured data from the crowd by presenting crowdworkers with a spreadsheet like interface where workers can enter missing values. This however limits the information that can be provided to the crowdworkers related to the dataset and attribute relations.

\section{Conclusions, limitations and future work}\label{sec:conc}
We conclude our study by answering the questions we asked in the introduction:
\begin{enumerate}
    \item Can humans fill in (impute) missing qualitative and quantitative data?\\
    Our results show that indeed, humans can be effectively used to fill in missing qualitative and quantitative data.
    \item If the same missing value is imputed multiple times by humans, will it have similar variation as in values imputed by machine based multiple imputation methods?\\
    Our results show that variation induced by multiple imputation using CrowdMI is similar to the variation induced by machine based imputations.
    \item How much information is needed by crowdworkers for imputing missing observations?\\
    We have shown that providing a dataset context with some descriptives about attribute correlations and plots for effective visualization provides an optimal amount of information needed by crowdworkers to fill in missing values.
\end{enumerate}

\textbf{Limitations and further work}: In this study, we have only used three datasets (two independent datasets, one perturbed version) to evaluate CrowdMI. This is partly due to increased costs with the use of more questionnaires and partly due to the fact that three datasets are enough to establish usefulness of a new method and to do a feasibility analysis. Which is inline with other evaluations done in crowdsourcing, some of which only use a single dataset. For our future work however, we would like to validate our method on more datasets of varying variety. We only considered relatively low dimensional datasets, our future attempts will focus on large dimensional datasets with information presented using a small subset of variables that are strongly associated with the attribute having missing values.

\bibliographystyle{aaai}

\bibliography{aaai}

\end{document}